\documentclass[letterpaper, 10 pt, journal, twoside]{IEEEtran}

\usepackage[ruled,vlined]{algorithm2e}
\usepackage{amsmath}
\usepackage{amssymb}
\usepackage{color}
\usepackage{graphicx}
\usepackage[hidelinks]{hyperref}
\usepackage{subcaption}

\newtheorem{definition}{Definition}
\SetKw{Break}{break}
\DeclareMathOperator\erf{erf}
\graphicspath{{figures/}}

\begin{document}

\title{Robotic Pick-and-Place With Uncertain Object Instance Segmentation and Shape Completion}

\author{Marcus Gualtieri and Robert Platt%
\thanks{The authors are with the Khoury College of Computer Sciences, Northeastern University, Boston, MA, USA (email {\tt\footnotesize mgualti@ccs.neu.edu}). This work was supported in part by NSF (1724257, 1724191, 1763878, 1750649) and NASA (80NSSC19K1474).}}

\maketitle

\begin{abstract}
We consider robotic pick-and-place of partially visible, novel objects, where goal placements are non-trivial, e.g., tightly packed into a bin. One approach is (a) use object instance segmentation and shape completion to model the objects and (b) use a regrasp planner to decide grasps and places displacing the models to their goals. However, it is critical for the planner to account for uncertainty in the perceived models, as object geometries in unobserved areas are just guesses. We account for perceptual uncertainty by incorporating it into the regrasp planner's cost function. We compare seven different costs. One of these, which uses neural networks to estimate probability of grasp and place stability, consistently outperforms uncertainty-unaware costs and evaluates faster than Monte Carlo sampling. On a real robot, the proposed cost results in successfully packing objects tightly into a bin 7.8\% more often versus the commonly used minimum-number-of-grasps cost.
\end{abstract}


\section{Introduction}

\IEEEPARstart{P}{ick-and-place} is prehensile manipulation where objects are grasped rigidly and placed into desired configurations \cite{Mason2018}. This problem has been extensively studied for fully observed objects, resulting in deeper understanding of the problem and efficient planning algorithms \cite{Tournassoud1987,Alami1991,Alami1994,Nielsen2000,Krontiris2015,Wan2019}. Such a planner could be combined with a separately designed perceptual algorithm for estimating objects' geometry from raw sensor data. However, a system with separately designed perception and planning modules is not always optimal: these methods treat grasping an unobserved part of an object the same as grasping a part that is fully observed, which could lead to avoidable failures.

One approach to this problem is to dispense with the idea of separate perception and planning modules and use reinforcement learning (RL) to train a single module that does both. While some success has been achieved with this idea \cite{Gualtieri2018A,Gualtieri2018B,Gualtieri2020}, training is time-consuming, the system is not robust to changes in either task or environment, and performance is often suboptimal, even for simple tasks (cf. placing mugs with an RL approach \cite{Gualtieri2018A} versus a modular approach \cite{Manuelli2019}).

Another approach is to plan in belief space, i.e., in probability distributions over state \cite{Kaelbling1998}. While this handles arbitrary types of uncertainty, there are a couple of important drawbacks. First, these methods often require a detailed description of the observation and state transition models of the system, which can be difficult to obtain \cite{Kaelbling2013,Xiao2019}. Second, planning takes place in the space of probability distributions over states, which is continuous and, for practical problems, high dimensional. For these reasons, this approach has been confined to problems with few dimensions or other simplifying structure.

We take a new approach to pick-and-place of novel, partially visible objects: (a) use perception to predict the complete geometry of the objects and (b) incorporate instance segmentation and shape completion uncertainty as a planning cost. We compare seven cost functions, four of which explicitly model the probability of successfully executing a regrasp plan, including grasp quality (GQ), Monte-Carlo (MC) sampling, uncertainty at contact points (CU), and success prediction (SP). With only small modifications to existing planners, we efficiently account for perceptual uncertainty.


We test this approach with bin packing and bottle arrangement tasks in both simulation and the real world. Results show perception is indeed a significant source of error and shape completion is critical to regrasp planning. Also, the SP method consistently outperforms three other methods (no cost, step cost, and GQ) which do not account for perceptual uncertainty in terms of avoiding grasp failures.  Furthermore, the SP cost is faster than MC sampling.

\section{Related Work}

\textbf{Pick-and-place in fully observed environments:} Pick-and-place was often studied independently from perception. \textit{Regrasping}, which is to find a sequence of picks and places moving an object to a goal pose, was first explained by Tournassoud et al. \cite{Tournassoud1987}. There is a discrete search component, for sequencing grasp-place combinations, and a continuous search component, for connecting grasp-place combinations with a motion plan. Alami et al. generalized regrasping to multiple, movable objects, pointed out the problem is NP-hard, and coined the term \textit{manipulation planning} \cite{Alami1991}. Later they considered different cost functions for the discrete search, including path length and number of grasp changes \cite{Alami1994}. Nielsen and Kavraki gave a 2-level, probabilistically complete planner for manipulation planning \cite{Nielsen2000}. Wan et al. employed a 3-level planner, where the high-level planner provides a set of goal poses for the objects, the middle-level planner is a regrasp planner, and the low-level planner is a motion planner \cite{Wan2019}. For \textit{non-monotonic} rearrangement problems (i.e., objects need moved more than once), a middle-level planner displacing multiple objects was more efficient \cite{Krontiris2015}. Our approach is to start with a well-established regrasp (i.e., middle-level) planner and build an uncertainty capability upon it.


\textbf{Pick-and-place of known objects:} Others have considered pick-and-place of imperfectly perceived objects with known shapes. One approach is to match object models to sensor data, as in Tremblay et al. \cite{Tremblay2018}. Morgan et al. used clustering to localize blocks for their box and blocks benchmark task \cite{Morgan2019}. However, we consider novel objects, i.e., the shapes are not known \textit{a priori}.

\textbf{Pick-and-place of novel objects:} A few projects have considered novel-object pick-and-place, where the complete shapes of objects are not given. The first to address this was Jiang et al. \cite{Jiang2012A} who used random sampling with classification to identify placements that are likely to be stable and satisfy human preference. After this, we approached the problem with deep RL by learning a grasp/place value function \cite{Gualtieri2018A,Gualtieri2018B,Gualtieri2020}. Next, Manuelli et al. proposed a 4-component pipeline: (a) instance segmentation, (b) key point detection, (c) optimization-based planning for task-specific object displacements, and (d) grasp detection \cite{Manuelli2019}. Objects were minimally represented by key points, which are 3D points indicating task-relevant object parts, e.g., the top, bottom, and handle of a mug. Later, Gao and Tedrake augmented this with shape completion, which is useful for avoiding collisions when planning arm motions with the held object \cite{Gao2019}. Finally, Mitash et al. addressed the problem by fusing multiple sensor views and allowing a single regrasp as necessary, conservatively assuming the object is as large as its unobserved region \cite{Mitash2020}. None of these compared different ways of accounting for perceptual uncertainty, as we do here.


\textbf{Pick-and-place under uncertainty:} A general approach to pick-and-place under arbitrary types of uncertainty is to solve a partially observable Markov decision process (POMDP). Kaelbling and Lozano-P{\'e}rez focused on symbolic planning in belief space with black-box geometric planners and state estimators \cite{Kaelbling2013}. Xiao et al. used POMCP \cite{Silver2010} to update their belief about the arrangement of a small set of known objects \cite{Xiao2019}. However, the POMDP approach requires significant computation and an accurate model of transition and sensor dynamics.

\textbf{Grasping under uncertainty:} We extend ideas from grasping under object shape uncertainty to pick-and-place planning. The two most common approaches to grasping under shape uncertainty are (a) evaluate grasp success over an MC sampling of object shapes and (b) evaluate a probabilistic model of grasp success. Kehoe et al. took the MC approach and represented uncertainty as normally distributed polygonal vertices with given means and variances \cite{Kehoe2012A}. Hsiao et al. provided a probabilistic model for grasp success given multiple object detections and grasp quality evaluations \cite{Hsiao2011}. Afterward, Gaussian process implicit surfaces (GPISs) were proposed as a representation of object shape uncertainty \cite{Dragiev2011,Mahler2015,Laskey2015,Li2016}. GPISs combine multiple observations of an object's signed distance function (SDF) into a Gaussian process -- a normal distribution over SDFs \cite{Dragiev2011}. Mahler et al. compared a probabilistic model (based on the variance of the GPIS at contact points) to an MC approach \cite{Mahler2015}. The MC approach did better but has higher computational cost. Laskey et al. improved the efficiency of MC sampling from the GPIS by employing multi-armed bandit techniques to reduce the number of evaluations for grasps that are unlikely to succeed \cite{Laskey2015}. Li et al. conducted real-world experiments filtering grasps with different thresholds on variance of the GPIS at contact points \cite{Li2016}. Lundell et al. represented objects as voxels, used a deep network to complete objects, and performed MC sampling using dropout \cite{Lundell2019}.





\section{Problem Statement}
\label{sec:problem}

Consider planning robot motions to place a partially visible object of unknown shape into a goal pose. In particular, we consider this problem in the context of the following system:

\begin{definition}[Move-open-close system]
A \emph{move-open-close system} consists of one or more objects, a robotic manipulator, and one or more depth sensors, each situated in 3D Euclidean space. Objects are rigid masses $O_1, \dots, O_{n_\mathit{obj}} \subseteq \mathbb{R}^3$, sampled randomly from an unknown probability distribution. The manipulator is equipped with a parallel-jaw gripper with status \textit{empty} or \textit{holding}. The action of the robot is to move the gripper to a target pose $T_e \in \mathit{SE}(3)$, followed by either gripper \textit{open} or \textit{close}. At each step, the robot acquires a point cloud $C \in \mathbb{R}^{n\times3}$, observes its gripper status, and takes an action.
\end{definition}

To simplify planning, we avoid dynamic actions (e.g., pushing). In particular, \textit{close} actions should fix an object rigidly in the gripper, and \emph{open} actions should place an object at rest. Assume an \textit{antipodal grasp} is sufficient to fix an object in the gripper (\cite{Murray1994} p. 233), and assume the conditions in \cite{Tournassoud1987} are sufficient to stably place an object on a horizontal surface. We now state the problem as follows:

\begin{definition}[Regrasping under perceptual uncertainty]
Given a move-open-close system, objects represented as point clouds $\{\bar{C}_i \in \mathbb{R}^{\bar{n}_i \times 3}\}_{i=1}^{n_\mathit{obj}}$, perceptual uncertainty vectors $\{\mathcal{U}_i \in \mathbb{R}^{d_i}\}_{i=1}^{n_\mathit{obj}}$, and a set of goal poses for each object $\{\{T_{ij} \in \mathit{SE}(3)\}_{j = i}^{n_\mathit{goal}}\}_{i = 1}^{n_\mathit{obj}}$, find a sequence of antipodal grasps and stable places maximizing the probability of displacing an object to a goal pose.
\label{def:regraspingUnknownObjects}
\end{definition}


There are different ways to represent the uncertainty vectors $\mathcal{U}_i$, including point-wise segmentation/completion uncertainties, Monte-Carlo samples, and grasp/place success prediction networks: these are described in Section~\ref{sec:regrasp}. Intuitively, actions should account for uncertainty in object shape, as grasping and placing on uncertain object parts is likely to result in unpredictable movements of the object.

\section{System Overview}
\label{sec:system}

\begin{figure}[ht]
  \centering
  \includegraphics[width=0.5\textwidth,trim={2.77cm 7.75cm 5.85cm 8.85cm},clip]{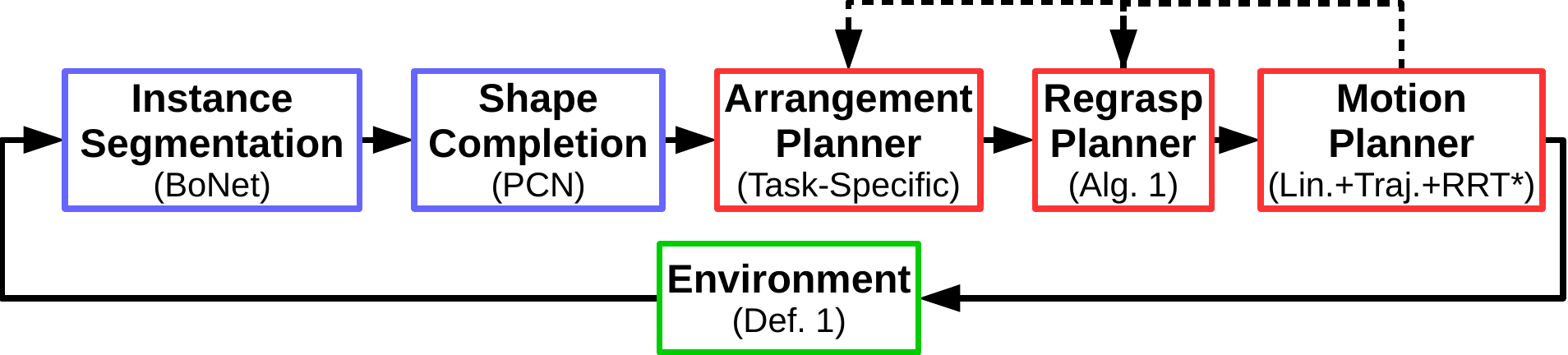}
  \caption{Diagram of our system architecture. Green represents the environment, blue the perceptual modules, and red the planning modules. Dashed arrows are followed up to a number of times if no plan is found.}
  \label{fig:architecture}
\end{figure}

Consider a modular, perception-planning pipeline for displacing partially visible, novel objects, where the regrasp planner addresses the problem of Def.~\ref{def:regraspingUnknownObjects}. Such a system is summarized in Fig.~\ref{fig:architecture}. For each perception-action cycle, the environment produces a point cloud, the geometry of the scene is estimated, a partial plan for displacing an object is found, and the first pick-and-place of the plan is executed. Automatic resensing and replanning accounts for failures, similar to MPC \cite{Morari1999}. In this section, each component is briefly described. Regrasping under segmentation and completion uncertainty -- the main contribution -- is detailed in Section~\ref{sec:regrasp}.

\subsection{Perception}
\label{sec:perception}

The purpose of the perceptual modules is to reconstruct the geometry of the scene so we can apply geometric planning algorithms. Additionally, they must quantify their own uncertainty so plans unlikely to succeed can be avoided. For both instance segmentation and shape completion, we have chosen point clouds as the input/output representation of objects. A point representation consumes less memory than uncompressed voxel grids, enables efficient planning, and, from our previous experience, exhibits good simulation-to-real domain transfer \cite{Gualtieri2018A,Gualtieri2018B,Gualtieri2020}.

\subsubsection{Object instance segmentation}

The input to the segmentation module is a point cloud $C \in \mathbb{R}^{n\times3}$, and the output is a point cloud for each object, $\{C_i \in \mathbb{R}^{n_i\times3}\}_{i=1}^{n_\mathit{obj}}$ with $\sum_{i=1}^{n_\mathit{obj}} n_i \leq n$, and uncertainties $\{U_i \in \mathbb{R}^{n_i}\}_{i=1}^{n_\mathit{obj}}$. Although any object instance segmentation method with this interface can be used in the proposed architecture, our implementation uses BoNet \cite{Yang2019}. BoNet produces an $n \times K$ matrix, where $K$ is a predefined maximum number of objects, and each row is a point's distribution over object ID. $U_i$ is the $\max$ of the $i$th row, which is the estimated probability the $i$th point is correctly segmented. (And, optionally, points with $U_i$ below a threshold can be omitted.)


\subsubsection{Shape completion}

\begin{figure}[ht]
  \centering
  \begin{subfigure}{0.325\columnwidth}
    \centering
    \includegraphics[width=\textwidth]{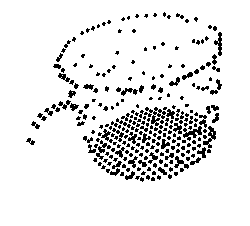}\\
    \includegraphics[width=\textwidth]{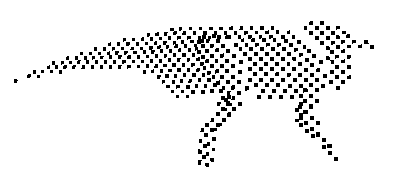}
    \caption{Observed cloud.}
    \label{fig:completion-a}
  \end{subfigure}
  \begin{subfigure}{0.325\columnwidth}
    \centering
    \includegraphics[width=\textwidth]{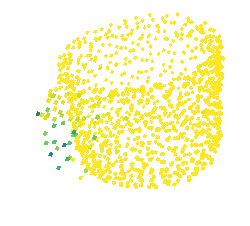}\\
    \includegraphics[width=\textwidth]{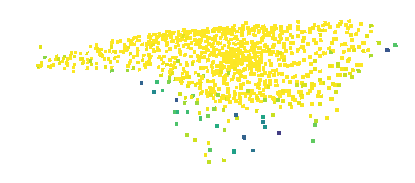}
    \caption{Completed cloud.}
    \label{fig:completion-b}
  \end{subfigure}
  \begin{subfigure}{0.325\columnwidth}
    \centering
    \includegraphics[width=\textwidth]{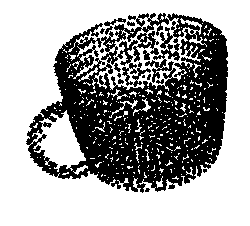}\\
    \includegraphics[width=\textwidth]{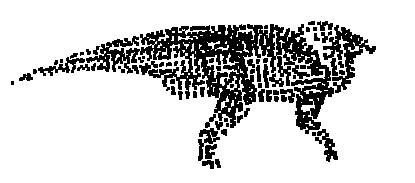}
    \caption{Ground truth.}
    \label{fig:completion-c}
  \end{subfigure}
  \caption{Shape completions with PCN. Yellow represents high $\bar{U}$ values (near 1), and blue represents low $\bar{U}$ values (near 0.5).}
  \label{fig:completion}
\end{figure}

The input to the shape completion module is a point cloud $C \in \mathbb{R}^{n\times3}$, and the output is a point cloud $\bar{C} \in \mathbb{R}^{\bar{n}\times3}$ that is a dense sampling of points on all object faces, including faces not visible to the sensor. We also require an uncertainty estimate for each completed point, $\bar{U} \in \mathbb{R}^{\bar{n}}$. Although any shape completion method with this interface can be used in the proposed architecture, our implementation uses a modified version of PCN \cite{Yuan2018}. PCN consists of an encoder (two PointNet layers \cite{Qi2017A}) and a decoder (three fully connected, inner product layers). We augmented the original version of PCN with a second decoder for uncertainty estimates. In particular, the uncertainty decoder is trained using a binary cross-entropy loss to predict the probability each point is within Euclidean distance $\beta \in \mathbb{R}_{++}$ of the nearest ground truth point. So the uncertainty values should be interpreted as the estimated probability each completed point is accurate. Example completions are shown in Fig.~\ref{fig:completion-b}.




\subsection{Planning}

We use a 3-level planner, similar to Wan et al. \cite{Wan2019}.

\subsubsection{Arrangement planner}

The input to the arrangement planner is a list of completed clouds, $\bar{C}_1, \dots, \bar{C}_{n_\mathit{obj}}$, and the output is a set of triples $\{(T, c, i)_j\}_{j=1}^{n_\mathit{goal}}$, where $T$ is a goal pose for object $i$ and $c$ is an associated goal cost. The reason the arrangement planner produces multiple goals for multiple objects is to increase the chances one of them is feasible. Besides, not all goals are equal: some may be more preferable to the task. For example, in bin packing, some placements will result in tighter packings than others. This is captured by the goal cost, $c$. We implement a different arrangement planner for each task.

\subsubsection{Regrasp planner}

The regrasp planner takes in the triples from the arrangement planner and produces a sequence of picks and places, i.e., effector poses, that displaces one object. If a regrasp plan is not found, more goals can be requested from the arrangement planner (as indicated by dashed lines in Fig.~\ref{fig:architecture}).

\subsubsection{Motion planner}

The motion planner finds a continuous motion between picks and places. Any off-the-shelf motion planner will do: we use a 3-level planner that first attempts a linear motion, then Trajopt \cite{Schulman2013}, and then RRT* with timeout \cite{Karaman2011}. If no motion plan is found, the regrasp planner can be resumed from where it left off, but marking the infeasible section so the same solution is not found again.

\section{Regrasp Planning Under Uncertainty}
\label{sec:regrasp}

Regrasps are needed due to kinematic constraints: the grasps at the object's current pose may all be in collision or out of reach at the object's goal poses. In this case, a number of temporary places (i.e., non-goal places) are needed. Our regrasp planner (Alg.~\ref{alg:regrasp}) extends Tournassoud et al.'s \cite{Tournassoud1987} to handle multiple goals for multiple objects, arbitrary additive costs, and discrete grasp/place sampling. Related planners (e.g, \cite{Alami1991,Alami1994,Nielsen2000,Krontiris2015}) could also have been adapted to the purpose: the main point is to incorporate segmentation and shape completion uncertainty into the cost.

\begin{algorithm}[ht]
\DontPrintSemicolon
\SetKwInOut{input}{Input}
\SetKwFunction{SampleGrasps}{SampleGrasps}
\SetKwFunction{SampleTemporaryPlaces}{SampleTemporaryPlaces}
\SetKwFunction{UpdateRegraspGraph}{UpdateRegraspGraph}
\SetKwFunction{A}{A$^*$}
\input{Number of sampling iterations $N$, completed cloud $\bar{C}$, uncertainty vector $\mathcal{U}$, goal poses and costs $\{(T,c)_j\}_{j = 1}^{n_\mathit{goal}}$, and $\mathit{costLowerBound}$.}
$\mathit{RG} \gets \left[ \right]$\;
\For{$i \gets 1, \dots, N$}{
  $G, \mathit{gc} \gets \SampleGrasps(\bar{C}, \mathcal{U})$\;
  $P, \mathit{pc} \gets \SampleTemporaryPlaces(\bar{C}, \mathcal{U})$\;
  $\mathit{RG} \gets \UpdateRegraspGraph(\mathit{RG},$
  $\{(T,c)_j\}_{j=1}^{n_\mathit{goal}}, G, \mathit{gc}, P, \mathit{pc})$\;
  $\mathit{plan}, \mathit{cost} \gets \A(\mathit{RG})$\;
  \lIf{$\mathit{cost} \leq \mathit{costLowerBound}$} {\Break}
}
\Return $\mathit{plan}$
\caption{Regrasp planner: run for each object.}
\label{alg:regrasp}
\end{algorithm}

A key part of Alg.~\ref{alg:regrasp} is the regrasp graph, \textit{RG}. The \textit{regrasp graph} is a matrix where rows refer to grasps and columns refer to places. When the object has been grasped, column changes are allowed to switch the object's placement, and when the object has been placed, row changes are allowed to switch grasps \cite{Tournassoud1987}. To Tournassoud et al.'s regrasp graph we add costs: matrix values are the sum of the corresponding grasp and place costs if the grasp-place combination is feasible (i.e., there is a collision-free IK solution) and infinity otherwise.

Alg.~\ref{alg:regrasp} is run in parallel for each object that has at least one goal pose. For $N$ steps, additional grasps ($G$ with costs $\mathit{gc}$) and temporary places ($P$ with costs $\mathit{pc}$) are randomly sampled. Given the shape completion $\bar{C}$, grasp samples are constrained to satisfy the geometric antipodal conditions (\cite{Murray1994} p. 233), and place samples are constrained to satisfy the stability conditions (\cite{Tournassoud1987}). The function \texttt{UpdateRegraspGraph} adds a row for each sampled grasp and a column for each sampled place to \textit{RG} and then checks IK and collisions for the new grasp-place combinations. Finally, A* with a consistent heuristic finds an optimal pick-and-place sequence, given the current samples \cite{Hart1968}. Next we define the cost function used by A* and give different ways of calculating grasp and place costs.

\subsection{Maximize probability of regrasp plan execution success}

The aim is to choose a regrasp plan that maximizes the joint probability each grasp is antipodal and each temporary place is stable, i.e., maximize Eq.~\ref{eq:pps1}, where $G_i$ is the event the $i$th grasp is antipodal, $P_i$ is the event the $i$th place is stable, and $m$ is total number of picks and places. Assuming each grasp/place is independent of previous steps in the plan, we arrive at Eq.~\ref{eq:pps2}.\footnote{Assuming knowledge that a previous grasp/place was successful does not decrease the joint probability of success, Eq.~\ref{eq:pps2} is a lower bound.} Taking the log and abbreviating $\Pr(G_i)$ as $g_i$ and $\Pr(P_i)$ as $p_i$ yields Eq.~\ref{eq:pps3}.

\begin{align}
\begin{split}
&\Pr(G_1, P_1, \dots, G_{m/2})\\
&=\Pr(G_{\frac{m}{2}} | G_1, P_1, \dots, P_{\frac{m}{2} - 1}) \cdots \Pr(P_1 | G_1) \Pr(G_1)
\end{split} \label{eq:pps1} \\
&\approx \Pr(G_{m/2}) \cdots \Pr(P_1) \Pr(G_1) \label{eq:pps2}
\end{align}

\begin{equation}
\log\left[\Pr(G_1, \dots, G_{m/2})\right] \approx \sum_{i = 1}^{m/2} \log(g_i) + \sum_{i=1}^{m/2 - 1} \log(p_i) \label{eq:pps3}
\end{equation}

Negating Eq.~\ref{eq:pps3} results in a non-negative, additive cost: the form required by A*. We account for plan length and task cost by adding these as objectives to a multi-criterion optimization problem (\cite{Boyd2004} pp. 181-184). Scalarization results in Eq.~\ref{eq:regraspCost}, where $w_1, \dots, w_4 \in \mathbb{R}_{++}$ are trade-off parameters and $c \in \mathbb{R}$ is the task cost associated with the goal placement (from the arrangement planner). This is the cost used by our regrasp planner. To complete the description, we next look at different ways of estimating $g_i$ and $p_i$.

\begin{equation}
w_1 m - w_2 \sum_{i=1}^{m/2} \log(g_i) - w_3 \sum_{i=1}^{m/2 - 1} \log(p_i) + w_4 c \label{eq:regraspCost}
\end{equation}

\subsection{Probability grasps are antipodal and places are stable}
\label{sec:graspPlaceProb}

\subsubsection{Grasp quality (GQ)}

One way to estimate $g_i$ is via a measure of ``robustness'' of the grasp to small perturbations in the nominal shape completion. For antipodal grasps, Murray et al. suggest choosing grasps where the line between contacts is inside and maximally distant from the edges of both friction cones (\cite{Murray1994} p. 233). This way, a grasp will satisfy the geometric antipodal conditions under small perturbations to the object's shape.

We place this idea into our probabilistic framework. For both grasp contacts, $j = 1, 2$, let $\theta_j \in [0, \pi]$ be the angle between the surface normal $n_j$ and the normalized, outward-pointing vector $b_j$ connecting both contacts. Assume $\theta_j$ is distributed according to a truncated normal distribution with mode $\mu_j$ and scale $\sigma$, where the angle $\mu_j$ (Eq.~\ref{eq:gq-1}) is derived from the nominal object shape and $\sigma$ is given. The probability $b_j$ lies in the friction cone is then $\Pr(\theta_j \leq \theta_\mathit{max}) = F(\theta_\mathit{max};\mu_j,\sigma,0,\pi)$, where $F$ is the cumulative density function of the truncated normal distribution and $\theta_\mathit{max}$ is half the angle of the friction cone. We make the simplifying assumption that this probability is independent between contacts, giving Eq.~\ref{eq:gq-2}.

\begin{align}
\mu_j &= \arccos(b_j \cdot n_j) \label{eq:gq-1}\\
g_i &= \prod_{j=1}^2 F(\theta_\mathit{max}; \mu_j, \sigma, 0, \pi) \label{eq:gq-2}
\end{align}

The effect of the GQ estimator is to choose grasps that are as centered as possible in both friction cones, given the estimated object shape. The scale parameter $\sigma$ makes the trade-off between regrasp plan length and centering of grasps: small $\sigma$ prefers centered grasps over short plans and large $\sigma$ prefers short plans over centered grasps.

\subsubsection{Monte Carlo (MC)}

Another approach is to estimate $g_i$ and $p_i$ via segmentation and completion samples, as was done for grasping under shape uncertainty \cite{Kehoe2012A,Mahler2015,Laskey2015,Lundell2019}. The idea is to randomly generate multiple segmentations then completions and average grasp/place antipodal/stability.

Let $\Pr(\bar{\mathcal{C}}_i | C)$, for $i=1, \dots, n_\mathit{obj}$, be a distribution over object shapes, where $C$ is the input point cloud, from which we collect samples. This could be implemented with segmentation/completion networks with randomized components, e.g., using dropout \cite{Lundell2019}. However, to compare to the CU method (described next), we use the point-wise uncertainty outputs of the networks ($U_i$ and $\bar{U}_i$ in Section~\ref{sec:perception}) as follows.


For segmentation, the object ID for each point is independently sampled from the distributions given by the segmentation matrix. (To reduce noise, we only sample points whose $U$-value is below a threshold.) For shape completion, assume the $i$th point's offset from the nominal point is i.i.d. $\sim \mathcal{N}(0,\sigma_i^2)$. Since $\bar{U}_i$ is the estimated probability the point is offset no more than $\beta$, the standard deviation of the point's offset is derived from the Gaussian CDF as in Eq.~\ref{eq:mc}. To sample a shape: (a) sample a segmentation point-wise using the segmentation mask and (b) compute the shape completion given this segmentation. Then, for each point in the completion, (b.1) sample a direction uniformly at random and (b.2) sample an offset along this direction from a normal distribution with 0 mean and standard deviation given by Eq.~\ref{eq:mc}.


\begin{equation}
\sigma_i = \frac{\beta}{\sqrt{2}\erf^{-1}(\bar{U}_i)} \label{eq:mc}
\end{equation}

Regardless of implementation, $g_i$ is estimated as $\text{\#antipodal}/M$ and $p_i$ is estimated as  $\text{\#stable}/M$ where $M$ is the number of shape samples and \#antipodal is the number of shapes for which the $i$th grasp is antipodal and \#stable is the number of shapes for which the $i$th place is stable.

\subsubsection{Contact uncertainty (CU)}

Computing $g_i$ and $p_i$ with MC is computationally expensive if $M$ is large. This motivates considering uncertainty only at contact points. For instance, placing an object on its unseen, predicted geometry could likely be unstable, so we penalize grasps/places on uncertain object parts. The same idea is behind penalizing high-variance grasp contacts \cite{Mahler2015,Li2016}. 

\begin{figure}[ht]
  \centering
  \includegraphics[height=1.5in]{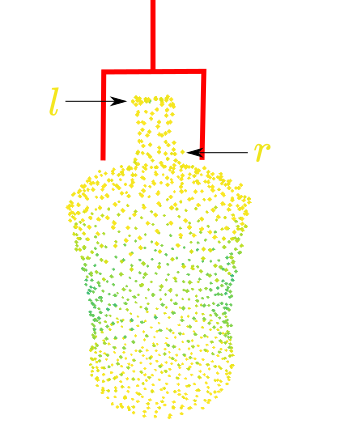}
  \includegraphics[height=1.5in]{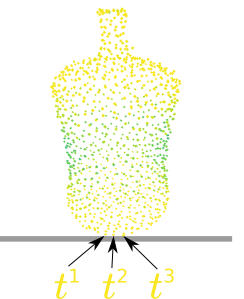}
  \caption{\textbf{Left}. For an antipodal grasp (shown in red), there are at least 2 contact points, $l$ and $r$. \textbf{Right}. For a stable placement on a flat surface, there are at least 3 contact points, $t^1$, $t^2$, and $t^3$. Colors represent estimates of $\Pr(\bar{V}_i | V_i)\Pr(V_i)$, where yellow represent higher probabilities.}
  \label{fig:contacts}
\end{figure}

Formally, suppose for the $i$th point, the segmentation network estimates $\Pr(V_i)$, where $V_i$ is the event the $i$th point is segmented correctly. Suppose the shape completion network estimates $\Pr(\bar{V}_i | V_i)$ where $\bar{V}_i$ is the event the $i$th point is within Euclidean distance $\beta$ of a ground truth point. Assuming whether a grasp (place) is antipodal (stable) depends only on whether each contact point is correctly segmented and is within Euclidean distance $\beta$ of the nearest ground truth point, and assuming independence between contacts, $g_i$ and $p_i$ are estimated via Eq.~\ref{eq:cu1}~and~\ref{eq:cu2}, where contacts are explained in Fig.~\ref{fig:contacts}.

\begin{align}
g_i &\approx \Pr(\bar{V}_l | V_l) \Pr(V_l) \Pr(\bar{V}_r | V_r) \Pr(V_r) \label{eq:cu1}\\
p_i &\approx \prod_{i=1}^3 \Pr(\bar{V}_{t^i} | V_{t^i}) \Pr(V_{t^i}) \label{eq:cu2}
\end{align}

The uncertainty values from PCN ($\bar{U}_i$ in Section~\ref{sec:perception}) are used to estimate $Pr(\bar{V}_i | V_i)$. Estimating $\Pr(V_i)$ from the uncertainty values from BoNet ($U_i$ in Section~\ref{sec:perception}) is less straight-forward since, for each completed point we must associate a corresponding segmentation uncertainty. A heuristic we found that works well for this is, for each point in the completed cloud, take the nearest neighbor in the segmented cloud.

\subsubsection{Success prediction (SP)}

$g_i$ and $p_i$ can also be directly estimated with a neural network. The encoding of grasp/place as input to the neural network is an important design choice that affects performance \cite{Gualtieri2016}. Here, we encode grasps as the points from the shape completion, $\bar{C}$, inside the gripper's closing region w.r.t. the gripper's reference frame  (cf. \cite{Liang2019}). For places, the completed cloud, $\bar{C}$, is transformed to the place pose and translated with the bottom-center of the cloud at the origin. For network architecture, we use PCN with a single output with sigmoid activation, trained with the binary cross-entropy loss. Training data is generated in simulation, so labeling ground truth antipodal/stable is straight-forward.

\section{Experiments}

We ran experiments in simulation and the real world to compare the different ways in Section~\ref{sec:graspPlaceProb} for accounting for object shape uncertainty in regrasp planning. We also compare to two baselines which do not account for object shape uncertainty: \textit{no cost}, which takes the first regrasp plan found, and \textit{step cost}, which includes the step cost term only ($w_1=1$ in Eq.~\ref{eq:regraspCost}). The step cost appears almost exclusively in the regrasping literature, e.g., \cite{Tournassoud1987,Alami1991,Alami1994,Wan2019}.


\subsection{Setup}

\begin{figure}[ht]
  \centering
  \includegraphics[height=1.65in]{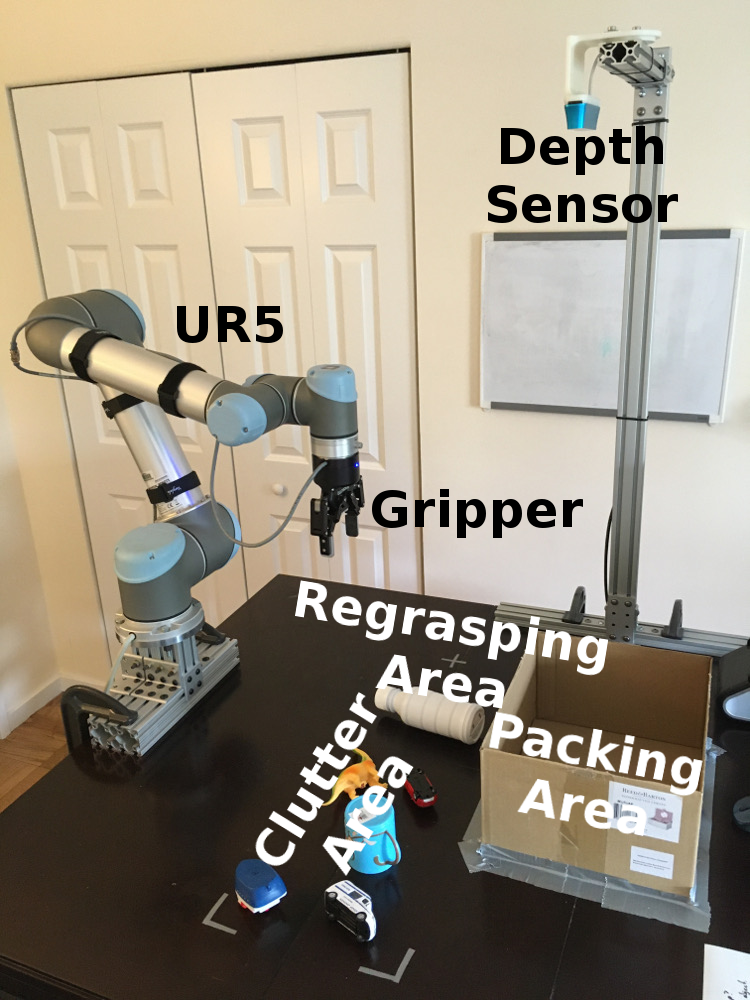} 
  \includegraphics[height=1.65in]{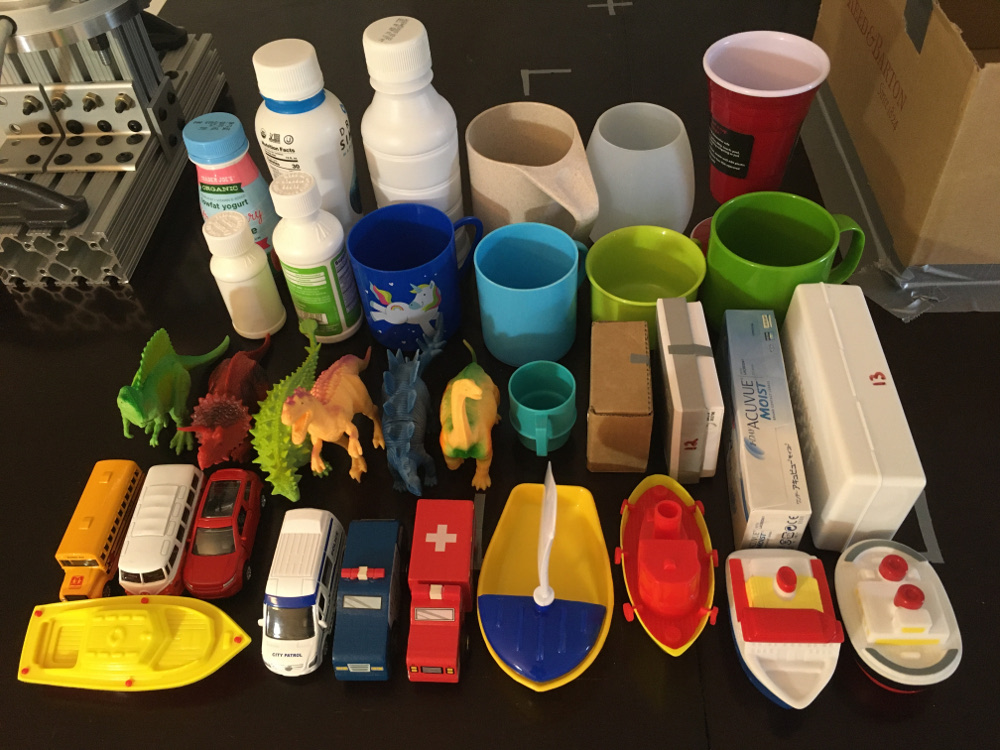} 
  \caption{\textbf{Left}. Environment includes a UR5 arm, a Robotiq 85 gripper, and a Structure depth sensor. \textbf{Right}. 34 same-category novel objects used for real-world packing experiments.}
  \label{fig:setup}
\end{figure}

The experimental environment is illustrated in Fig.~\ref{fig:setup}, left. We evaluate the proposed system on the following tasks:

\begin{enumerate}
\item \textbf{Bottle arrangement.} Place 2 bottles upright onto 2 coasters (from our prior work \cite{Gualtieri2018B,Gualtieri2020}).
\item \textbf{Bin packing.} Place 6 objects into a box minimizing packing height. This is known as the 3D irregular-shaped open dimension problem \cite{Wascher2007}. This is easier to evaluate than smallest bin size \cite{Wang2019} in the real world.
\item \textbf{Canonical placement.} Place any 1 of 5 objects into a goal pose. The arrangement planner is an oracle which consistently gives the same goal pose for an object. The purpose is to analyze regrasp performance independent from arrangement planner errors.
\end{enumerate}

\subsection{Simulation experiments}

The environment is simulated by OpenRAVE \cite{Diankov2010} using 3DNet objects \cite{Wohlkinger2012}. Objects are partitioned into 3 sets: \textit{Train} for training all deep networks, \textit{Test-1} for same-category novel objects (boat, bottle, box, car, dinosaur, mug, and wine glass), and \textit{Test-2} for novel-category objects (airplane, bowl, and stapler). A depth sensor, situated above the objects, captures a point cloud which is then passed into BoNet for segmentation. Grasps succeed if (a) exactly 1 object intersects the hand closing region, (b) the antipodal condition with $24^\circ$ friction cone is met, and (c) the robot is collision-free. Places are stable if the conditions in Tournassoud et al. \cite{Tournassoud1987} are met.


We evaluate \textit{place execution success rate} -- the proportion of regrasp plans with no grasp failures -- and \textit{temporary place stability rate} -- the proportion of temporary places that are stable. These metrics are fast to compute and indicate how well the regrasp plans succeed in placing objects. For the bin packing task we also have ``Packing height of 5'', which refers to the end-of-episode packing height when 5/6 objects are placed. This is a measure of how well the arrangement planner chooses goals. We use a 1-sided, same-variance, unpaired $t$-test to decide if one method significantly outperforms another. (If $p \leq 0.05$, we accept the hypothesis that the treatment outperforms the baseline.)

\subsubsection{Perception ablation study}

We quantify the potential benefit of accounting for uncertainty for the bin packing task. We evaluate performance with ground truth perception (GT Seg. \& Comp.), imperfect completion (GT Seg.), imperfect segmentation and completion (Percep.), and without shape completion (GT Seg. \& No Comp.) ``Imperfect'' means  the objects' segmentation/completion is estimated from the observed point cloud. Step and task costs are used, i.e., $w_1 = w_4 = 1$ and $w_2 = w_3 = 0$ in Eq.~\ref{eq:regraspCost}, where the task cost, $c$, is the estimated final packing height in centimeters.


\begin{table*}[h]
  \resizebox{\textwidth}{!}{%
  \centering
  \begin{tabular}{|l|c|c|c|c|c|c|}
  \hline
  & GT Seg. \& Comp. & GT Seg. (Train) & GT Seg. (Test-1) & Percep. (Train) & Percep. (Test-1) & GT Seg. \& No Comp.\\
  \hline
  Place Execution Success & 0.929 $\pm$ 0.008 & 0.767 $\pm$ 0.013 & 0.747 $\pm$ 0.013 & 0.718 $\pm$ 0.014 & 0.710 $\pm$ 0.014 & 0.508 $\pm$ 0.046\\
  \hline
  Regrasp Plan Found & 0.957 $\pm$ 0.006 & 0.882 $\pm$ 0.009 & 0.939 $\pm$ 0.007 & 0.879 $\pm$ 0.009 & 0.941 $\pm$ 0.007 & 0.100 $\pm$ 0.009\\
  \hline
  Grasp Antipodal & 0.931 $\pm$ 0.007 & 0.779 $\pm$ 0.013 & 0.761 $\pm$ 0.013 & 0.755 $\pm$ 0.013 & 0.736 $\pm$ 0.013 & 0.563 $\pm$ 0.047\\
  \hline
  Temporary Place Stable & 1.000 $\pm$ 0.000 & 0.769 $\pm$ 0.122 & 1.000 $\pm$ 0.000 & 0.828 $\pm$ 0.071 & 0.826 $\pm$ 0.081 & 0.500 $\pm$ 0.500\\
  \hline
  Packing height of 5 (cm) & 12.27 $\pm$ 0.315 & 12.36 $\pm$ 0.331 & 12.18 $\pm$ 0.306 & 12.37 $\pm$ 0.447 & 12.44 $\pm$ 0.307 & -- \\
  \hline
  Regrasp planning time (s) & 35.62 $\pm$ 1.103 & 38.46 $\pm$ 1.115 & 38.68 $\pm$ 1.141 & 35.76 $\pm$ 1.059 & 35.05 $\pm$ 1.077 & 15.86 $\pm$ 1.482\\
  \hline
  \end{tabular}}
  \caption{Perception ablation study for bin packing. Showing average $\pm$ standard error over 200 episodes.}
  \label{tab:perceptionAblation}
\end{table*}

Results (shown in Table~\ref{tab:perceptionAblation}) are as expected. A clear drop in performance is observed as perception becomes imperfect (down 18\% for imperfect completion and another 4\% for imperfect segmentation). Thus, a large source of error is due to perception, so there is space for improvement by accounting for perceptual uncertainty. Without shape completion, regrasp planning is crippled (regrasp plan found rate drops from 94.1\% for Percep. Test-1 to just 10.0\%). This is because insufficient grasp and place samples are found to displace objects.


\subsubsection{Regrasp cost comparison}

We test the hypothesis that a method estimating perceptual uncertainty (either MC, CU, or SP from Section~\ref{sec:graspPlaceProb}) selects regrasp plans that execute successfully more often on average than other methods (e.g., no cost, step cost, and GQ). Additional methods can be obtained by combinations, e.g., ``MC+GQ'' refers to the case where MC and GQ costs are summed.


Results for bin packing are shown in Table~\ref{tab:costComparisonPacking}. For Test-1, MC+GQ has the best grasp performance while SP has the best temporary place stability rate. MC+GQ significantly outperforms GQ ($p=0.0092$ for place execution success and $p=0.0073$ for grasp antipodal), suggesting the network's uncertainty estimates are useful for planning. For Test-2, SP has the best grasp performance (vs. step cost, $p=0.023$ for place execution success and $p=0.0023$ for grasp antipodal), and MC+GQ has the best temporary place stability rate. CU does not significantly outperform the baselines for either dataset: it is not sufficient to account for uncertainty only at the contact points. For bin packing, we do not see a significant improvement for place stability over the step cost, but this is because regrasps are rare with the step cost, obscuring the significance of the results. 


\begin{table*}[ht]
  \resizebox{\textwidth}{!}{%
  \centering
  \begin{tabular}{|l|c|c|c|c|c|c|c|}
  \hline
  & No Cost & Step Cost & GQ & MC & MC + GQ & CU & SP\\
  \hline
  Place Execution Success & 0.651 $\pm$ 0.013 & 0.725 $\pm$ 0.012 & 0.748 $\pm$ 0.012 & 0.756 $\pm$ 0.012 & \textbf{0.787} $\pm$ 0.011 & 0.712 $\pm$ 0.013 & 0.779 $\pm$ 0.012\\
  \hline
  Grasp Antipodal & 0.737 $\pm$ 0.011 & 0.751 $\pm$ 0.012 & 0.794 $\pm$ 0.011 & 0.811 $\pm$ 0.011 & \textbf{0.830} $\pm$ 0.010 & 0.743 $\pm$ 0.012 & 0.823 $\pm$ 0.010\\
  \hline
  Temporary Place Stable & 0.784 $\pm$ 0.024 & 0.857 $\pm$ 0.097 & 0.845 $\pm$ 0.030 & 0.904 $\pm$ 0.028 & 0.883 $\pm$ 0.031 & 0.848 $\pm$ 0.054 & \textbf{0.959} $\pm$ 0.018\\
  \hline
  Plan Length & 2.665 $\pm$ 0.031 & \textbf{2.038} $\pm$ 0.008 & 2.293 $\pm$ 0.021 & 2.222 $\pm$ 0.019 & 2.201 $\pm$ 0.018 & 2.105 $\pm$ 0.013 & 2.233 $\pm$ 0.019\\
  \hline
  Regrasp planning time (s) & \textbf{4.904} $\pm$ 0.230 & 7.201 $\pm$ 0.393 & 84.56 $\pm$ 0.827 & 90.10 $\pm$ 0.892 & 126.5 $\pm$ 1.029 & 72.00 $\pm$ 0.835 & 86.61 $\pm$ 1.040\\
  \hline
  \hline
  Place Execution Success & 0.412 $\pm$ 0.017 & 0.417 $\pm$ 0.017 & 0.395 $\pm$ 0.017 & 0.458 $\pm$ 0.017 & 0.422 $\pm$ 0.017 & 0.429 $\pm$ 0.017 & \textbf{0.465} $\pm$ 0.017\\
  \hline
  Grasp Antipodal & 0.484 $\pm$ 0.017 & 0.449 $\pm$ 0.017 & 0.450 $\pm$ 0.017 & 0.504 $\pm$ 0.017 & 0.472 $\pm$ 0.017 & 0.457 $\pm$ 0.017 & \textbf{0.518} $\pm$ 0.017\\
  \hline
  Temporary Place Stable & 0.704 $\pm$ 0.051 & 0.714 $\pm$ 0.125 & 0.533 $\pm$ 0.075 & 0.750 $\pm$ 0.083 & \textbf{0.800} $\pm$ 0.082 & 0.778 $\pm$ 0.101 & 0.686 $\pm$ 0.080\\
  \hline
  Plan Length & 2.514 $\pm$ 0.036 & \textbf{2.094} $\pm$ 0.015 & 2.247 $\pm$ 0.024 & 2.167 $\pm$ 0.020 & 2.150 $\pm$ 0.019 & 2.118 $\pm$ 0.017 & 2.193 $\pm$ 0.022\\
  \hline
  Regrasp planning time (s) & \textbf{6.030} $\pm$ 0.237 & 8.484 $\pm$ 0.408 & 51.61 $\pm$ 1.113 & 58.56 $\pm$ 1.064 & 71.38 $\pm$ 1.333 & 50.92 $\pm$ 1.177 & 53.35 $\pm$ 1.159\\
  \hline
  \end{tabular}}
  \caption{Cost comparison for bin packing for (\textbf{top}) Test-1 (230 episodes) and (\textbf{bottom}) Test-2 (200 episodes).}
  \label{tab:costComparisonPacking}
\end{table*}

We see a bigger difference with the canonical task on Test-1 (Table~\ref{tab:costComparisonCanonical}). In this case, MC, CU, and SP methods have significantly higher temporary place stability rates than no cost (which happened to do better than step cost) ($p=0.010$, $0.005$, and $2.4\times10^{-10}$, respectively). There is no doubt SP outperforms GQ for place execution success rate ($p=9.7\times10^{-9}$) and for grasp antipodal rate ($p=3.8\times10^{-9}$).


For both packing and canonical tasks, the SP method does significantly better than the baselines or GQ in terms of place execution success (packing Test-1 $p\leq0.032$, packing Test-2 $p\leq0.013$, canonical Test-1 $p\leq9.7\times10^{-9}$, and canonical Test-2 $p\leq2.9\times10^{-4}$), which supports the hypothesis.

\begin{table*}[ht]
  \resizebox{\textwidth}{!}{%
  \centering
  \begin{tabular}{|l|c|c|c|c|c|c|c|c|}
  \hline
  & No Cost & Step Cost & GQ & MC & MC + GQ & CU & SP\\
  \hline
  Place Execution Success & 0.727 $\pm$ 0.010 & 0.777 $\pm$ 0.009 & 0.856 $\pm$ 0.008 & 0.852 $\pm$ 0.008 & 0.861 $\pm$ 0.008 & 0.830 $\pm$ 0.008 & \textbf{0.913} $\pm$ 0.006\\
  \hline
  Grasp Antipodal & 0.833 $\pm$ 0.007 & 0.824 $\pm$ 0.009 & 0.906 $\pm$ 0.006 & 0.902 $\pm$ 0.006 & 0.908 $\pm$ 0.006 & 0.857 $\pm$ 0.008 & \textbf{0.951} $\pm$ 0.005\\
  \hline
  Temporary Place Stable & 0.785 $\pm$ 0.015 & 0.623 $\pm$ 0.067 & 0.700 $\pm$ 0.031 & 0.852 $\pm$ 0.022 & 0.784 $\pm$ 0.030 & 0.885 $\pm$ 0.029 & \textbf{0.967} $\pm$ 0.012\\
  \hline
  Plan Length & 3.061 $\pm$ 0.029 & \textbf{2.079} $\pm$ 0.009 & 2.273 $\pm$ 0.016 & 2.286 $\pm$ 0.016 & 2.220 $\pm$ 0.014 & 2.157 $\pm$ 0.013 & 2.239 $\pm$ 0.015\\
  \hline
  Regrasp planning time (s) & \textbf{2.462} $\pm$ 0.061 & 6.413 $\pm$ 0.353 & 62.19 $\pm$ 0.326 & 117.6 $\pm$ 0.724 & 121.1 $\pm$ 0.577 & 54.88 $\pm$ 0.366 & 61.54 $\pm$ 0.900\\
  \hline
  \hline
  Place Execution Success & 0.446 $\pm$ 0.011 & 0.535 $\pm$ 0.012 & 0.520 $\pm$ 0.012 & 0.543 $\pm$ 0.012 & 0.566 $\pm$ 0.012 & 0.533 $\pm$ 0.012 & \textbf{0.591} $\pm$ 0.011\\
  \hline
  Grasp Antipodal & 0.585 $\pm$ 0.010 & 0.592 $\pm$ 0.011 & 0.612 $\pm$ 0.011 & 0.630 $\pm$ 0.011 & 0.650 $\pm$ 0.011 & 0.590 $\pm$ 0.011 & \textbf{0.674} $\pm$ 0.010\\
  \hline
  Temporary Place Stable & 0.690 $\pm$ 0.021 & 0.555 $\pm$ 0.046 & 0.608 $\pm$ 0.030 & 0.717 $\pm$ 0.032 & 0.621 $\pm$ 0.034 & 0.671 $\pm$ 0.036 & \textbf{0.742} $\pm$ 0.027\\
  \hline
  Plan Length & 3.265 $\pm$ 0.035 & \textbf{2.323} $\pm$ 0.018 & 2.686 $\pm$ 0.025 & 2.501 $\pm$ 0.022 & 2.474 $\pm$ 0.021 & 2.419 $\pm$ 0.020 & 2.518 $\pm$ 0.023\\
  \hline
  Regrasp planning time (s) & \textbf{4.278} $\pm$ 0.156 & 14.84 $\pm$ 0.539 & 68.87 $\pm$ 0.657 & 99.36 $\pm$ 0.818 & 99.02 $\pm$ 0.819 & 60.05 $\pm$ 0.633 & 74.08 $\pm$ 0.732\\
  \hline
  \end{tabular}}
  \caption{Cost comparison for canonical task for (\textbf{top}) Test-1 and (\textbf{bottom}) Test-2 over $2,000$ episodes.}
  \label{tab:costComparisonCanonical}
\end{table*}

\subsection{Real world experiments}

We seek to (a) see if the perceptual components, trained with simulated data, work well with real sensor data and (b) verify the importance of uncertainty seen in simulation results. For these experiments, same-category novel objects are used (Fig.~\ref{fig:setup}, right).

\begin{figure*}[h]
  \centering
  \includegraphics[width=0.076\textwidth]{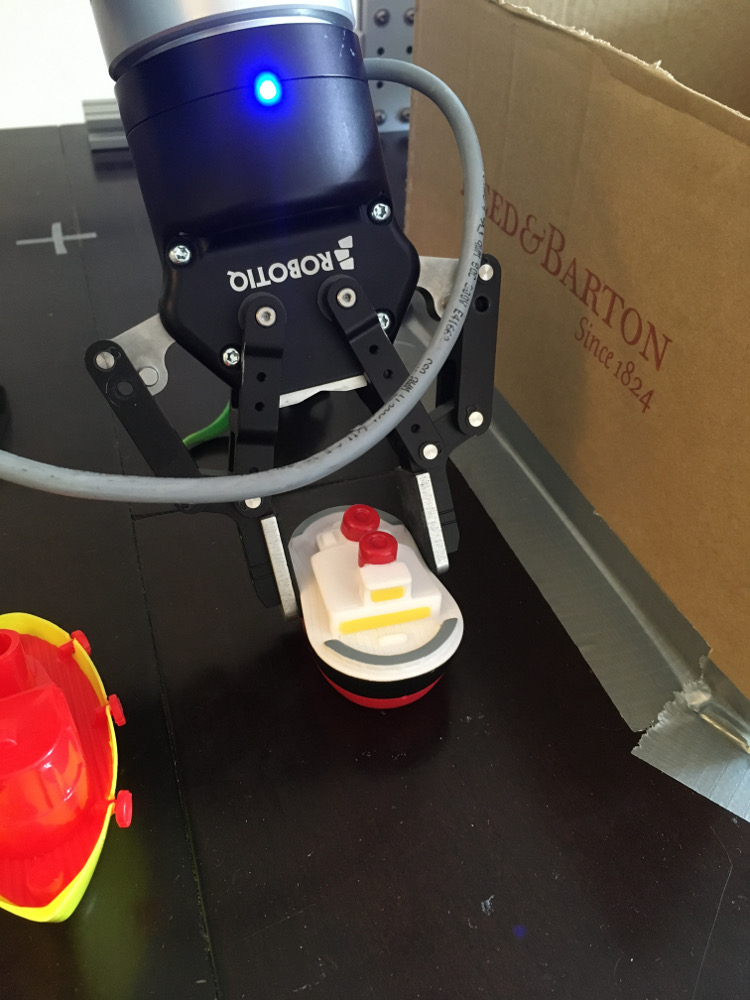}
  \includegraphics[width=0.076\textwidth]{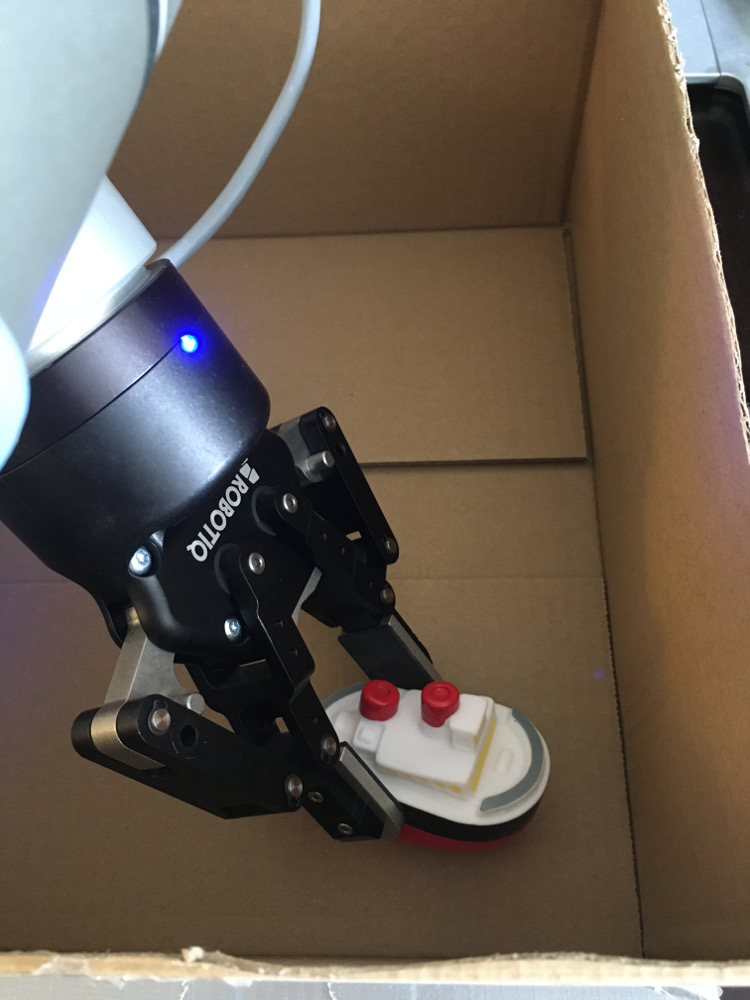}
  \includegraphics[width=0.076\textwidth]{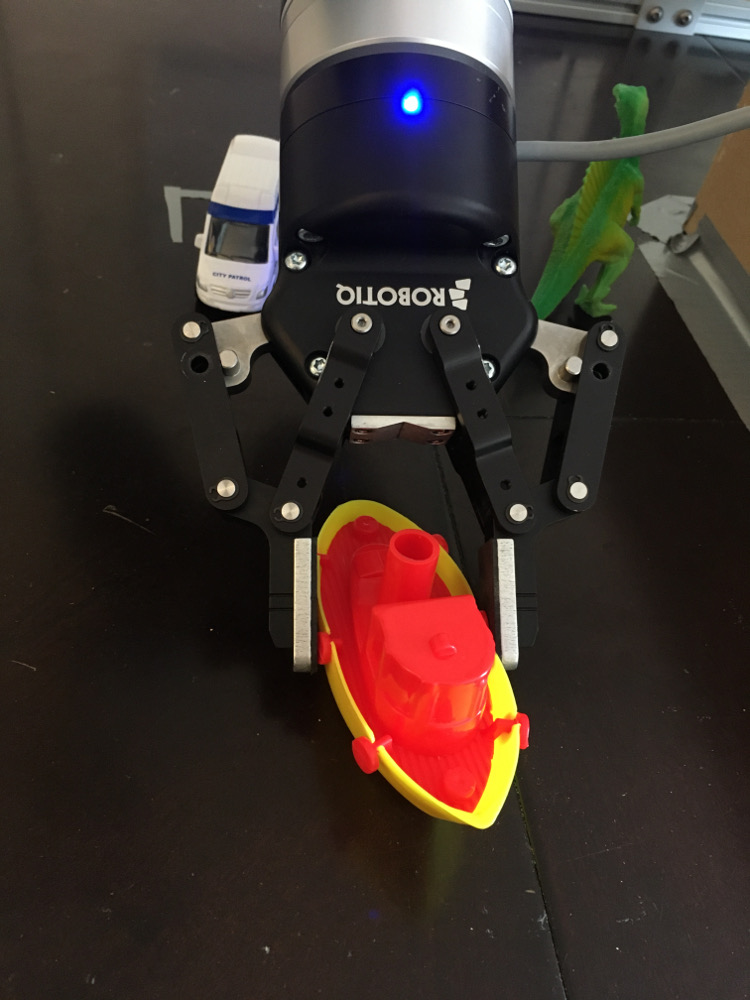}
  \includegraphics[width=0.076\textwidth]{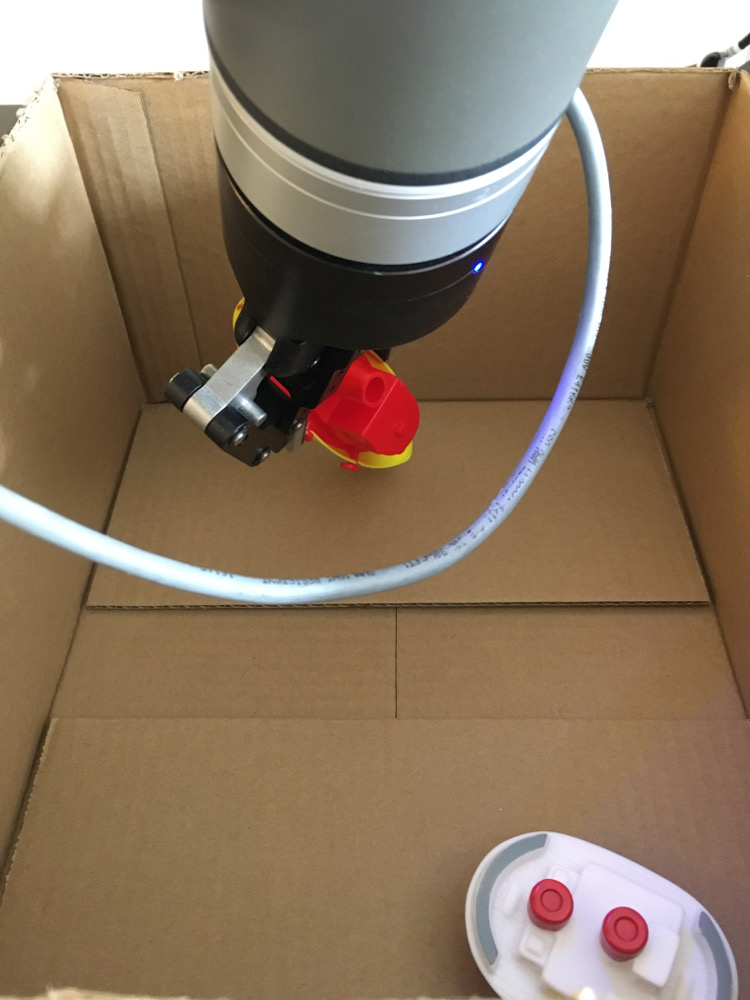}
  \includegraphics[width=0.076\textwidth]{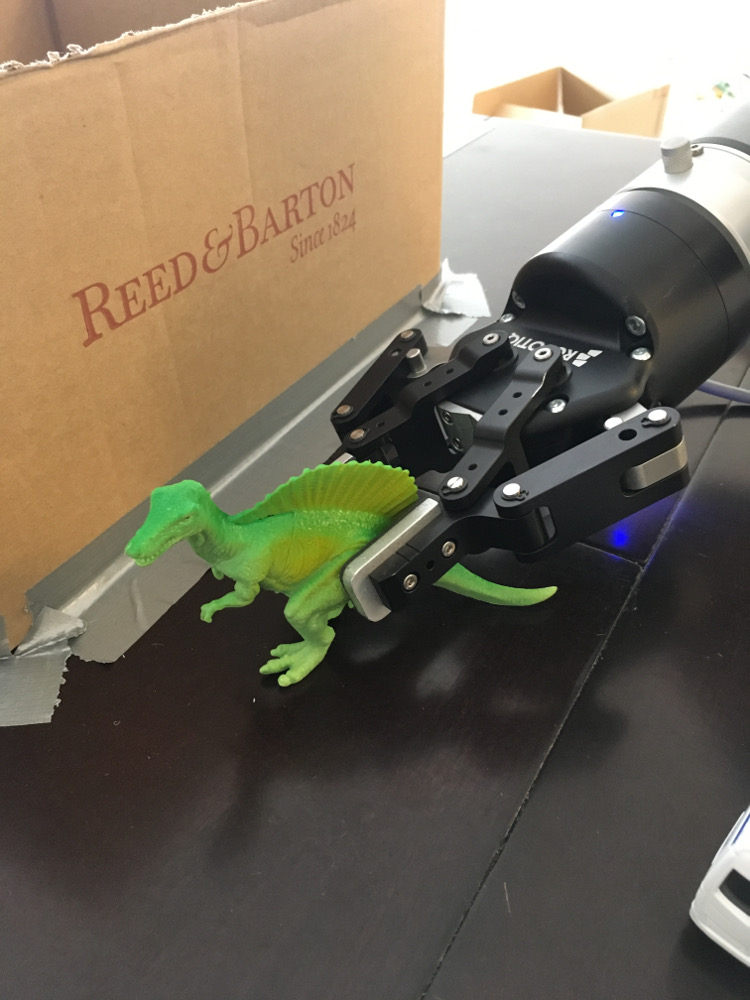}
  \includegraphics[width=0.076\textwidth]{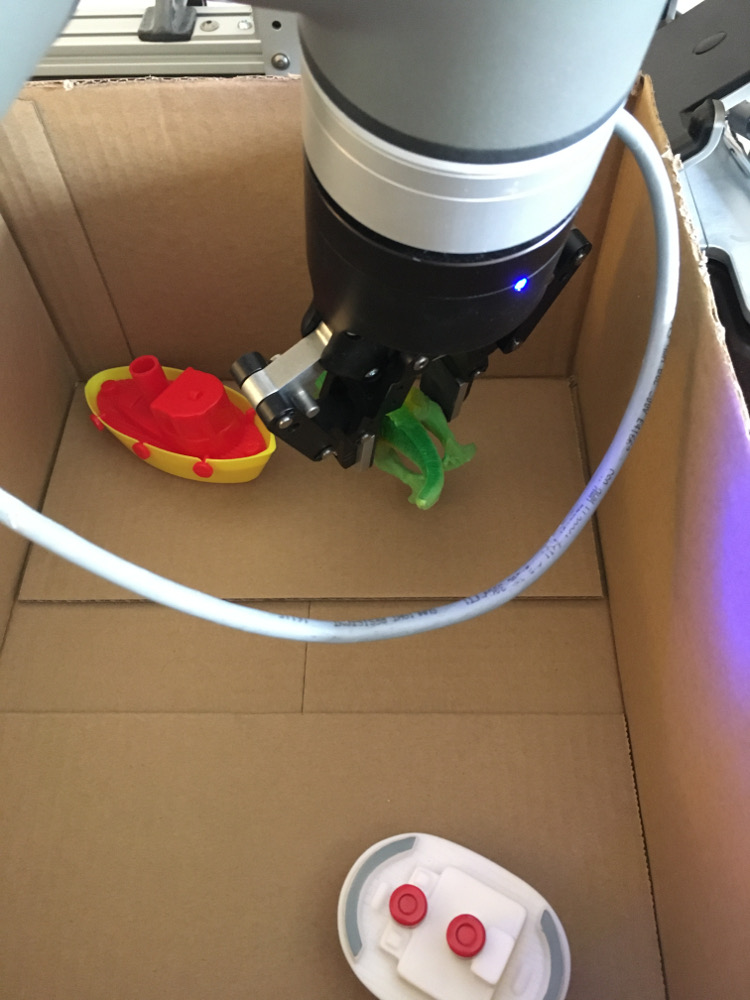}
  \includegraphics[width=0.076\textwidth]{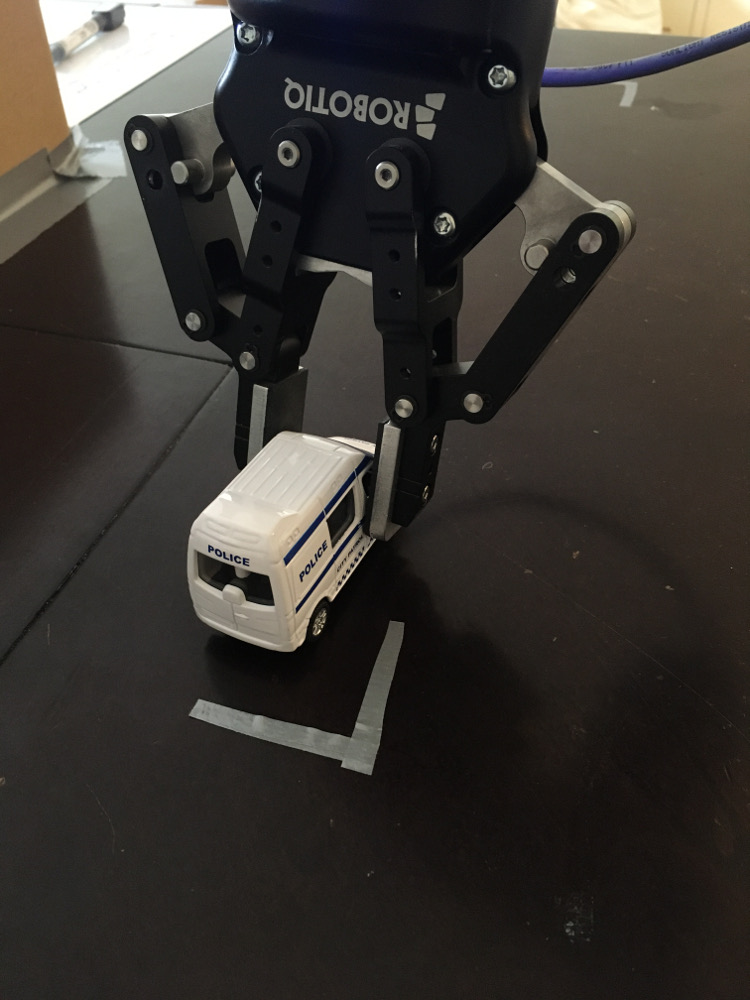}
  \includegraphics[width=0.076\textwidth]{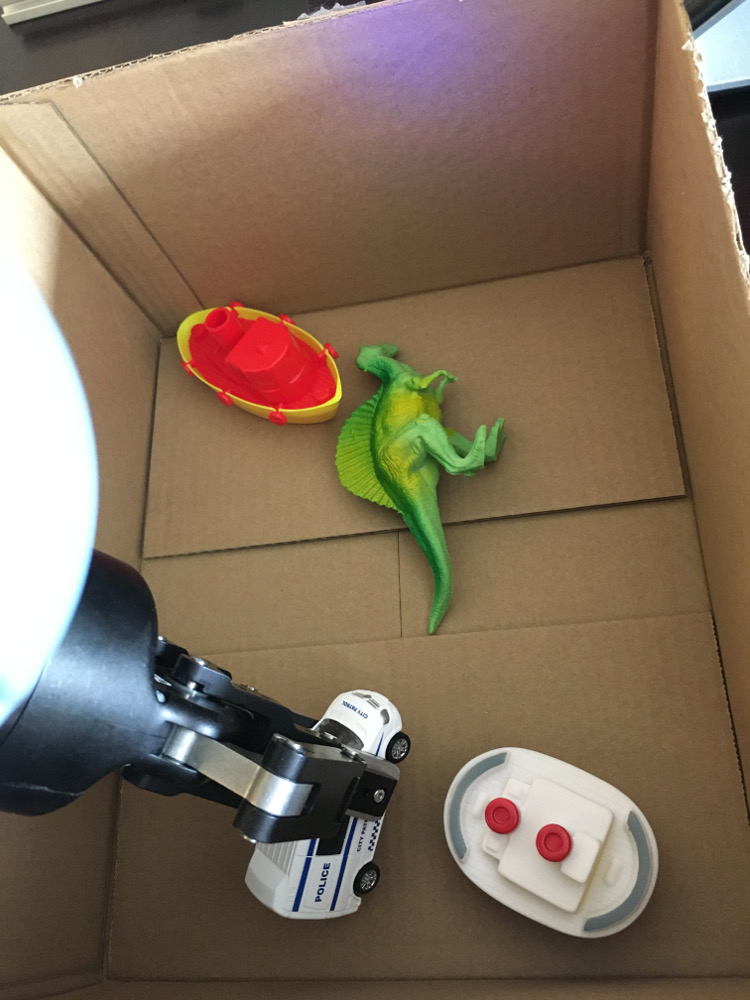}
  \includegraphics[width=0.076\textwidth]{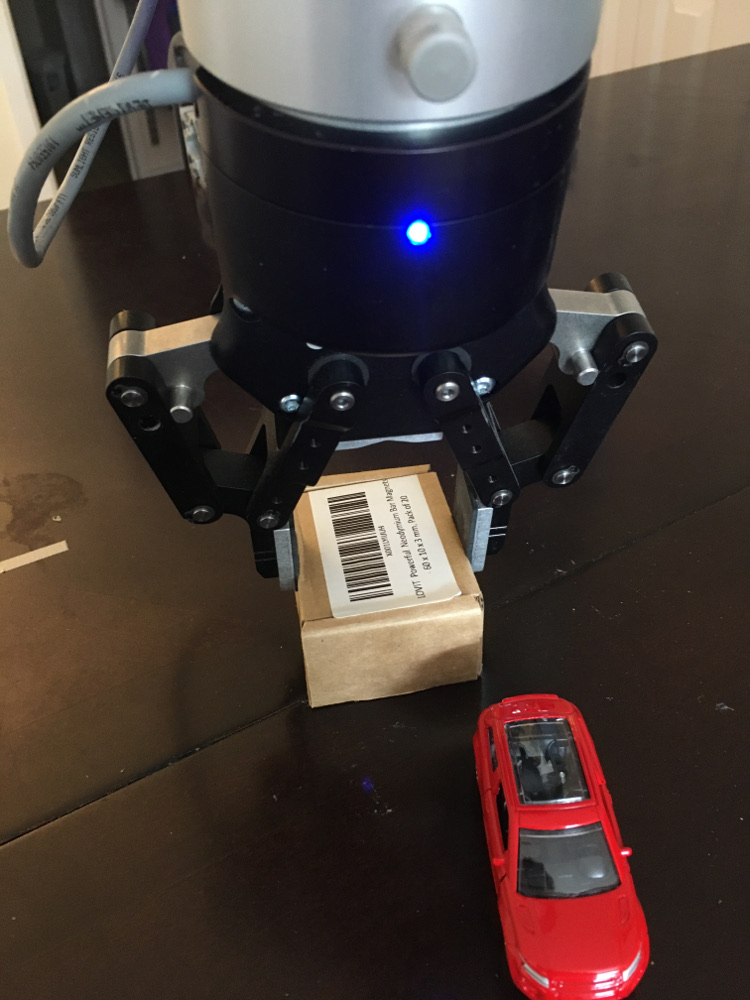}
  \includegraphics[width=0.076\textwidth]{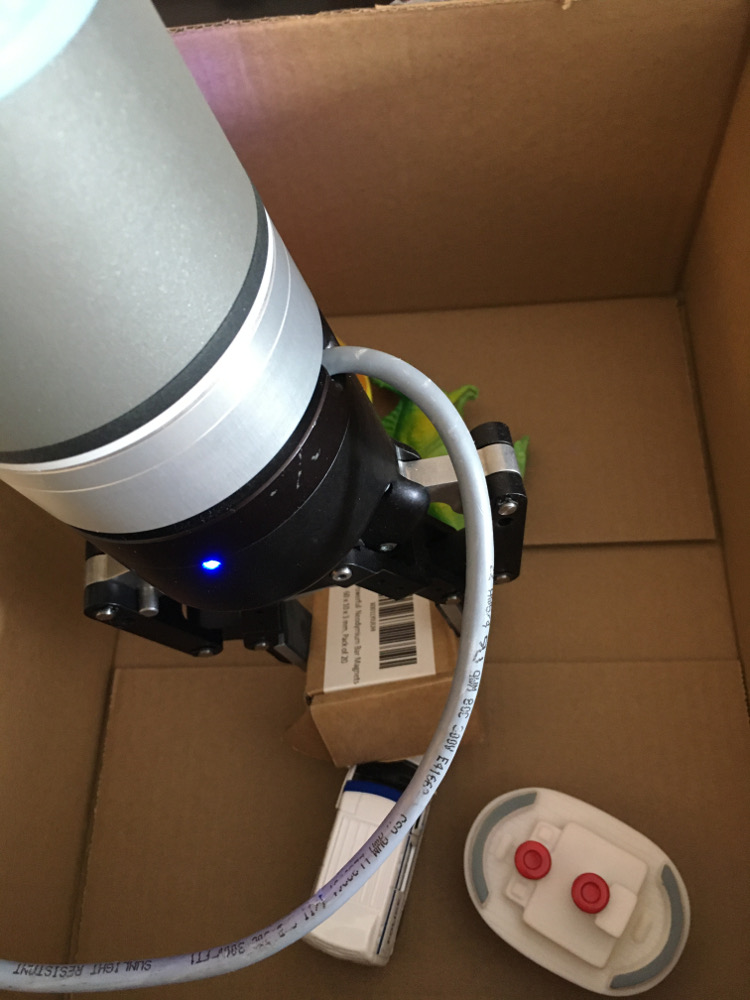}
  \includegraphics[width=0.076\textwidth]{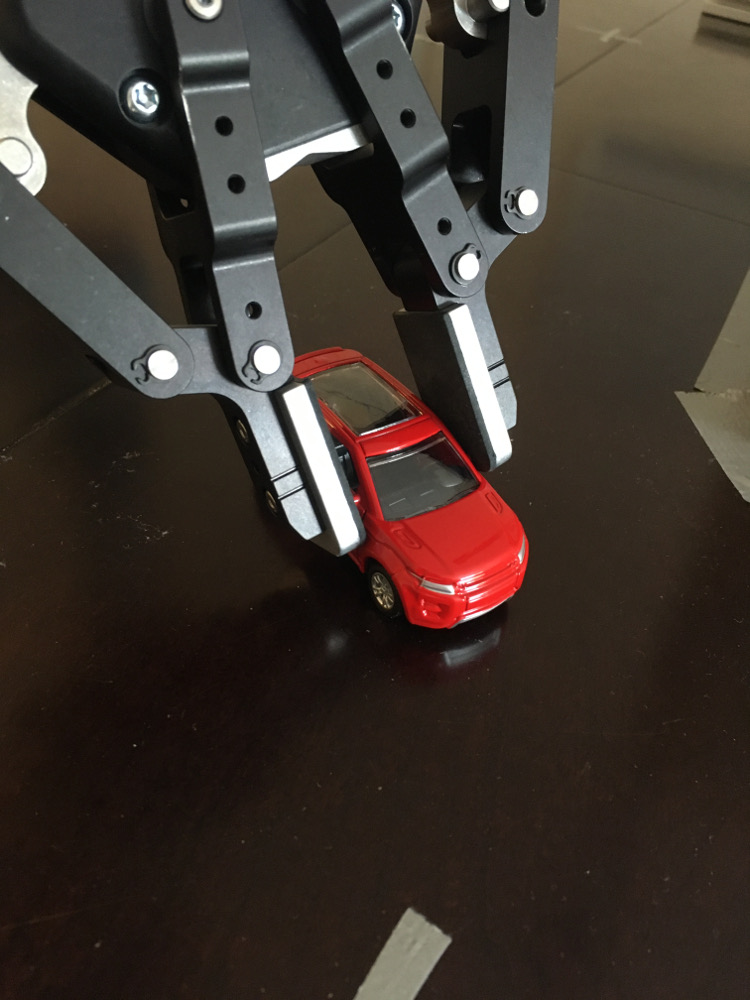}
  \includegraphics[width=0.076\textwidth]{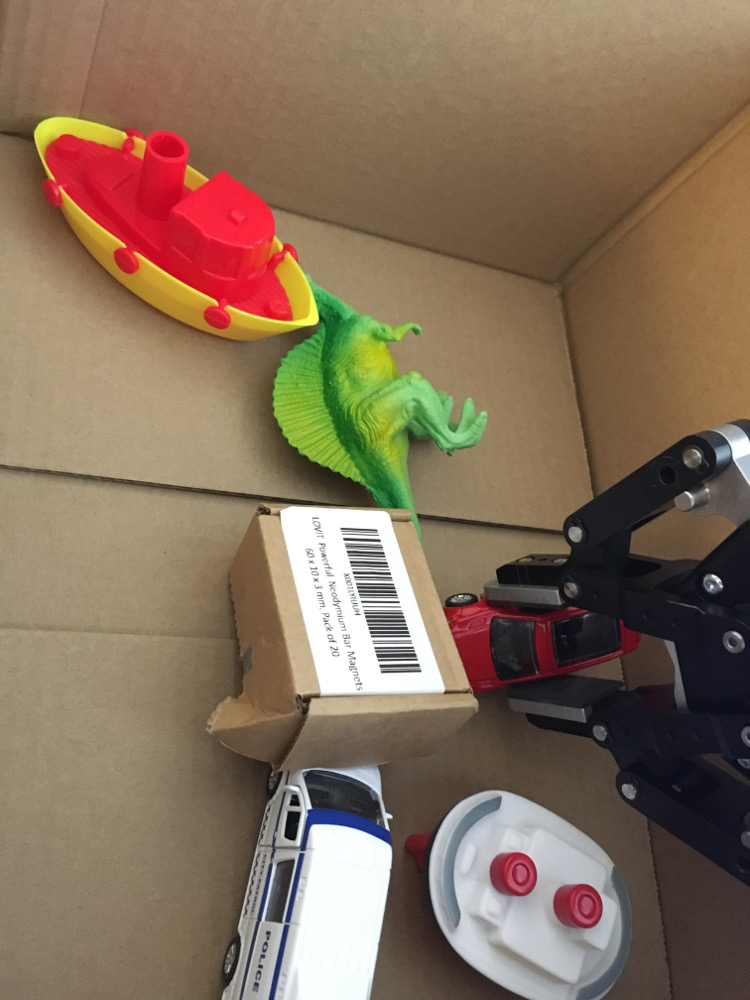}\\
  \smallskip
  \includegraphics[width=0.24178\textwidth]{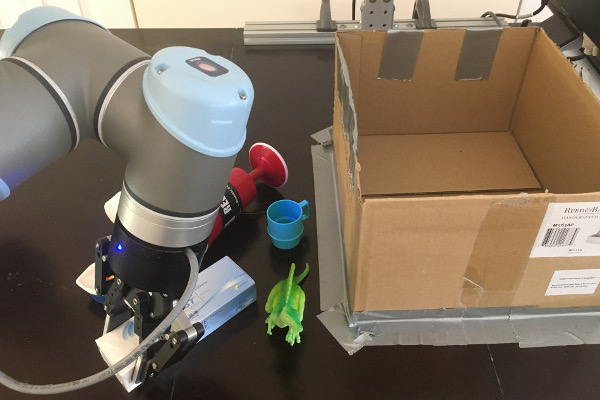}
  \includegraphics[width=0.24178\textwidth]{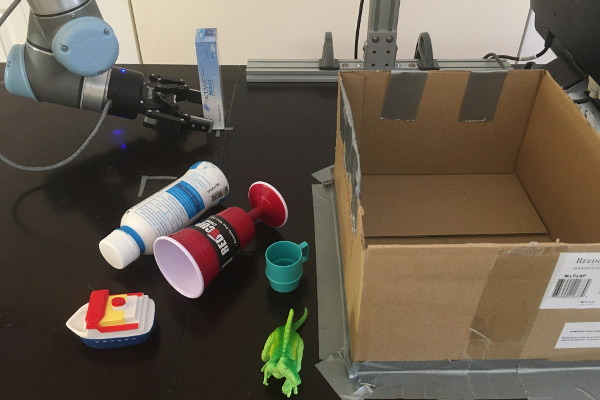}
  \includegraphics[width=0.24178\textwidth]{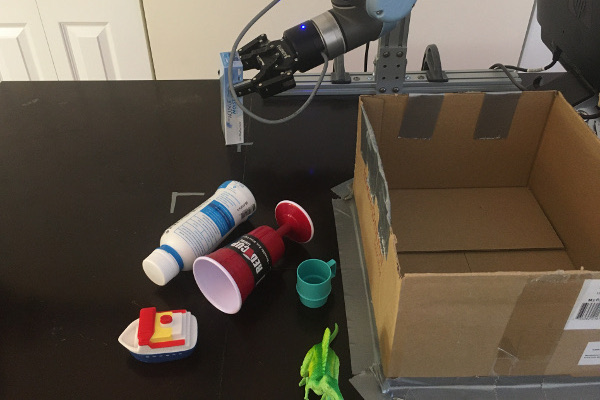}
  \includegraphics[width=0.24178\textwidth]{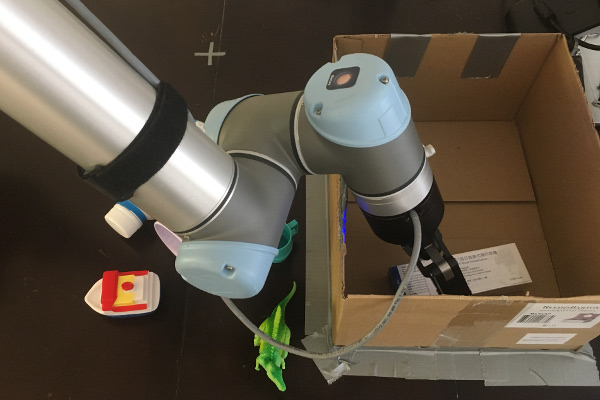}
  \caption{\textbf{Top}. Example packing sequence. \textbf{Bottom}. Example situation requiring a regrasp.}
  \label{fig:robotExperiments}
\end{figure*}

To answer part (a), no domain transfer was needed for bin packing. For bottles, BoNet (but not PCN) overfit to simulation data. This problem was mitigated by adding simulated sensor noise. To answer part (b), a regrasp cost comparison for bin packing is shown in Table~\ref{tab:robotPacking}. Both MC and SP methods significantly outperform the step cost (which outperforms GQ) ($p = 0.019$ and $p = 0.012$, respectively). Example packing and regrasp sequences are shown in Fig.~\ref{fig:robotExperiments}.

\begin{table}[h]
  \centering
  \resizebox{\columnwidth}{!}{%
  \begin{tabular}{|l|c|c|c|c|}
  \hline
  & Step Cost & GQ & MC & SP\\
  \hline
  Place Success Rate & 0.839 $\pm$ 0.027 & 0.833 $\pm$ 0.028 & 0.911 $\pm$ 0.021 & \textbf{0.917} $\pm$ 0.021\\
  \hline
  Grasp Success Rate & 0.883 $\pm$ 0.023 & 0.866 $\pm$ 0.024 & \textbf{0.947} $\pm$ 0.016 & 0.933 $\pm$ 0.017\\
  \hline
  Grasp Attempts & 196 & 201 & 207 & 210\\
  \hline
  Number of Regrasps & \textbf{17} & 21 & 27 & 30\\
  \hline
  Packing height of 5 (cm) & 7.333 $\pm$ 0.858 & \textbf{7.050} $\pm$ 0.650 & 7.588 $\pm$ 1.132 & 7.711 $\pm$ 0.880\\
  \hline
  \end{tabular}}
  \caption{Packing performance on the real robot. Showing average $\pm$ standard error over 30 episodes, each with 6 objects.}
  \label{tab:robotPacking}
\end{table}

We also compare bottle arrangement performance to our previous method, which uses RL to learn a pick-and-place policy \cite{Gualtieri2020}. Many of the same bottles as before are included, but 4/15 of them are more challenging. Two of the bottles are difficult to distinguish orientation (size of tops near size of bottoms), and two are near the 8.5 cm gripper width. Results are shown in Table~\ref{tab:robotBottles}. With the proposed method, all places are correct. Only the grasp success rate is lower than before, but all 3 grasp failures are with the wider bottles. Overall, we conclude the modular approach performs better (80\% vs. 67\% task success rate).\footnote{Source code and additional results are available at \url{https://github.com/mgualti/GeomPickPlace}.}

\begin{table}[h]
  \centering
  \begin{tabular}{|l|c|c|}
  \hline
  & Shape Completion & HSA \cite{Gualtieri2020}\\
  \hline
  Number of Objects Placed & 1.800 $\pm$ 0.074 & 1.667 $\pm$ 0.088\\
  \hline
  Task Success Rate & 0.800 $\pm$ 0.074 & 0.667 $\pm$ 0.088\\
  \hline
  Grasp Success Rate & 0.948 $\pm$ 0.029 & 0.983 $\pm$ 0.017\\
  \hline
  Place Success Rate & 1.000 $\pm$ 0.000 & 0.900 $\pm$ 0.040\\
  \hline
  \end{tabular}
  \caption{Bottles performance for the proposed method versus \cite{Gualtieri2020}. Showing average $\pm$ standard error over 30 episodes.}
  \label{tab:robotBottles}
\end{table}

\section{Conclusion}

Object instance segmentation and shape completion enable use of existing planning algorithms for pick-and-place of sensed objects. However, perceptual errors are still a major source of failure. To compensate for this, we compare different planning costs modeling probability of successfully executing a regrasp plan. Results show the SP cost, which uses separate networks to predict grasp/place success, consistently performs nearly as well as or outperforms all other costs. We attribute this to: (a) unlike baseline and GQ costs, SP can detect when perception is uncertain based on the distribution of perceived points; (b) unlike the CU cost, which considers uncertainty only at contact points, SP considers uncertainty at many points; and (c) unlike the MC cost, which requires sampling and evaluating multiple shapes, SP is computationally cheaper. On the other hand, when shape completion is accurate, e.g., when trained with one category like bottles, the step cost is a reasonable choice as planning and execution is faster than SP.

We note some limitations with our approach. First, the regrasp planner is much slower with a more sophisticated cost function than the step cost. This is because the step cost can exit the sampling loop when a two step plan is found, which occurs often in our experiments, while the other costs have no easy stopping criterion. Second, segmentation and completion accuracy is much lower with novel object categories. Third, integrating additional views to decrease uncertainty is an important aspect not considered.

\section*{Acknowledgements}

We thank Yuchen Xiao, Andreas ten Pas, Mary Gualtieri, Lawson Wong, Chris Amato, and the anonymous reviewers for their insightful feedback on early drafts.

\bibliographystyle{IEEEtran}
\bibliography{References}

\end{document}